\documentclass[11pt,letterpaper,onecolumn]{article}
\usepackage[left=2cm,top=1cm,right=3cm,nohead,nofoot]{geometry}
\usepackage[utf8]{inputenc}
\usepackage{hyperref}
\usepackage{amsmath}
\usepackage{amsfonts}
\usepackage{amssymb}
\usepackage{graphicx}
\usepackage{float}
\usepackage[affil-it]{authblk}
\begin{document}

\title{Time-series Scenario Forecasting}
\author{Sriharsha Veeramachaneni \thanks{Windlogics Inc., St. Paul, MN (Email: \texttt{hveera@gmail.com})}}
\date{\today}
\maketitle

\abstract{Many applications require the ability to judge uncertainty of time-series forecasts. Uncertainty is often specified as point-wise error bars around a mean or median forecast. Due to temporal dependencies, such a method obscures some information. We would ideally have a way to query the posterior probability of the \emph{entire} time-series given the predictive variables, or at a minimum, be able to draw samples from this distribution. We use a Bayesian dictionary learning algorithm to statistically generate an ensemble of forecasts. We show that the algorithm performs as well as a physics-based ensemble method for temperature forecasts for Houston. We conclude that the method shows promise for scenario forecasting where physics-based methods are absent.}

\section{Introduction}
Error bars (derived from conditional standard deviation) or predictive intervals (conditional quantiles) attempt to convey the uncertainty in a forecast to the user to enable more-informed decision making. Although such methods work well when forecasting univariate targets (e.g., total rainfall over the next week), they are insufficient for multivariate targets that have complex inter-dependencies (e.g., tomorrow's max temperature, max dew point and total rainfall). The basic problem is that the conditional errors of the forecast are not independent (e.g., if the forecast overestimates tomorrow's max temperature, it is likely to do the same for the dew point.) This is particularly true when forecasting time-series because most variables show strong temporal dependencies.

One approach taken for weather forecasting is to generate several \emph{scenarios} or realizations of the time-series, such that each realization satisfies the dependencies that are known the exist. If the scenario\footnote{We will use  ''scenario" for and individual forecast and ''event" for any property that can be computed from it.} forecasting method simulates draws from the posterior distribution, we may use it to answer complex queries about the forecast (e.g. compute the probability that there will be at least $1\,''$ of rain and max dew point is  $ < 80^\circ$,) which is impossible to do with simple error bars or predictive intervals on each of the variables\footnote{We might not be able to train a  system to forecast this particular event directly and use simple error bars because the events that interest the user may be diverse, change frequently or be unknown at training time.}.

Two major sources of weather forecast error are the uncertainty in the initial conditions and the incompleteness in the modeling of atmospheric physics. Based on this observation, weather scenario forecasting uses ensemble methods, where different physical models are used to make forecasts, each with several different of perturbations of the initial conditions~\cite{Wiki}. The perturbations need to be sensible in their ranges and statistical dependencies. Two commonly used methods to generate perturbations are based on \emph{SVD} (where the perturbations are generated in the direction of singular vectors, thereby accounting for correlations), and \emph{vector breeding} (where the perturbations are iteratively constructed in the most chaotic directions.) The NCEP Short-Range Ensemble Forecast (SREF) is one such product where 4 different models are run with several initial conditions~\cite{SREF}.

Although ensemble forecasting from multiple models allows the user to see a range of scenarios, it does not provide a way to accurately judge their likelihood. This is because there is no easy way to \emph{a priori} estimate the conditional probability that a particular physical model is the right one. This necessitates a statistical approach. Moreover, when forecasting variables for which there are no good physical models (e.g., wind turbine faults or electric load), a statistical approach may be the only recourse.

Dictionary learning is a statistical method to learn ''interesting" directions (or a basis) to compactly summarize high-dimensional data. Dictionary learning with sparse over-complete representations has been applied extensively in image and video processing~\cite{DictionaryTutorial}. 

We learn a dictionary jointly for the target time-series to be predicted and any available predictor variables, using a recently proposed Bayesian dictionary learning algorithm~\cite{bpfa}. At predict time, we draw samples from the conditional distribution of the target time-series given the observed predictors and the dictionary. Each of these samples is a scenario, and from this ensemble of scenarios the probability of any event of interest can be estimated.

\section{Statistical Scenario Generation}
Let us denote the $q$ dimensional target vector by $\bf{y}_i$ (e.g., the time-series of electric load over 84 hours starting at time $i$), and the corresponding $r$ dimensional predictor vector by $\bf{x}_i$ (e.g., the temperature, wind speed and dew point over the same 84 hours). We would like to construct a system to draw samples from the density $P(\bf{y}|\bf{x})$. Denote the concatenated vector $[\bf{x}_i, \bf{y}_i]$ by $\bf{z}_i$

We model $\bf{z}_i$ by the hierarchical model described in~\cite{bpfa}, which we repeat here for the sake of completeness.
\begin{eqnarray}
\bf{z}_i &=& \bf{D} \bf{w}_i + \bf{e}_i \nonumber \\
\bf{w}_i &=& \bf{b}_i \odot \bf{s}_i  \nonumber \\
\bf{d}_k &\sim & \mathcal{N}(0, 1/(q+r) \bf{I}) \nonumber \\
\bf{s}_i &\sim & \mathcal{N}(0, \gamma_s^{-1} \bf{I}) \nonumber \\
\bf{b}_{i,k} &\sim & \mbox{Bernoulli}(\pi_k) \nonumber \\
\pi_k &\sim & \mbox{Beta}(a/K, (K-1)b/K) \nonumber \\
\bf{e}_i &\sim & \mathcal{N}(0, \gamma_e^{-1} \bf{I}) \nonumber \\
\gamma_s &\sim & \mbox{Gamma}(c, d) \nonumber \\
\gamma_e &\sim & \mbox{Gamma}(e, f)
\end{eqnarray}
where $K$ is the number of dictionary atoms, $\bf{d}_k$ is the $k^{th}$ atom, $t$ is the dimension of $\bf{z}$, $\odot$ is the element-wise product, and $a, b, c, d, e, f$ are hyper-parameters. Note that the $(q+r) \times K$ dictionary matrix $\bf{D}$ can be partitioned into the $q \times K$ dictionary denoted $\bf{D}_y$ for the targets and the $r \times K$ dictionary denoted $\bf{D}_x$ for the predictors.

The parameters needed to generate data according to the model are given by $\Theta = (\bf{D}, \bf{\pi}, \gamma_s, \gamma_e)$. Given a training data, the model parameters can be estimated via Gibbs sampling (the formulae for which are derived in~\cite{bpfa}) yielding the estimate $\hat{\Theta}$.  

At predict time, we approximate $P(\bf{y}|\bf{x})$ as shown below. (In the equations below note that we are implicitly conditioning on the training data.)
\begin{eqnarray}
P(\bf{y}|\bf{x}) &=& \int_{\Theta, \bf{s}, \bf{b}} P(\bf{y}|\bf{x}, \Theta, \bf{s}, \bf{b}) P(\Theta, \bf{s}, \bf{b}| \bf{x}) \nonumber \\
&=&  \int_{\Theta, \bf{s}, \bf{b}} P(\bf{y}|\Theta, \bf{s}, \bf{b}) P(\Theta, \bf{s}, \bf{b}| \bf{x}) \nonumber \\
&=&  \int_{\Theta, \bf{s}, \bf{b}} P(\bf{y}|\Theta, \bf{s}, \bf{b}) P(\bf{s}, \bf{b}| \bf{x}, \Theta) P(\Theta | \bf{x}) \nonumber \\
&\approx & \int_{\bf{s}, \bf{b}} P(\bf{y}|\hat{\Theta}, \bf{s}, \bf{b}) P(\bf{s}, \bf{b}| \bf{x}, \hat{\Theta})  
\end{eqnarray}
where the second line is because the individual components of $\bf{z}$ are generated i.i.d., when $\Theta$, $\bf{s}$ and $\bf{b}$ are known, and for the fourth line we make the approximation $P(\Theta|\bf{x}) \approx \delta(\Theta - \hat{\Theta})$ (this is reasonable because the one test example will not alter the posterior distributions from those conditioned solely on the training data.)

Therefore in order to draw samples from the posterior distribution, we generate several samples of $\bf{s}$ and $\bf{b}$ from their posterior distribution given the estimated parameters and the predictor variables, and use each realization to reconstruct the target part of $\bf{z}$. The vectors $\bf{s}$ and $\bf{b}$ are drawn iteratively by Gibbs sampling just as was done during training. From given a sample $\bf{s}$ and $\bf{b}$ we generate a scenario for the targets as $\hat{\bf{y}} = \bf{D}_y (\bf{s} \otimes \bf{b})$. Note that the noise part of the target is not included in the realization.

\section{Experiments \& Discussion}
We evaluate our statistical scenario generation method by comparing to the SREF scenarios for the temperature in Houston. (We could have experimented with other variables such as electric load, but lacking physics-based scenarios as a baseline, the metrics would not have made intuitive sense.) Each instance of the target is the temperature over 84 hours starting at 03Z of a particular day, and each scenario is a 03Z temperature forecast extending out to 84 hours. We had training data for approximately 600 days of temperature, over which we performed 10 fold cross-validation for the statistical approach. We chose $K=100$ and a burn-in $=100$ for training and prediction.

The SREF ensemble comprises the 21 combinations of physical models and initial conditions: \emph{em.ctl}, \emph{em.n1}, \emph{em.n2}, \emph{em.p1}, \emph{em.p2}, \emph{eta.ctl1}, \emph{eta.ctl2}, \emph{eta.n1}, \emph{eta.n2}, \emph{eta.p1}, \emph{eta.p2}, \emph{nmm.ctl}, \emph{nmm.n1}, \emph{nmm.n2}, \emph{nmm.p1}, \emph{nmm.p2}, \emph{rsm.ctl1}, \emph{rsm.n1}, \emph{rsm.n2}, \emph{rsm.p1}, \emph{rsm.p2}. We chose the 4 models \emph{em.ctl}, \emph{eta.ctl1},  \emph{nmm.ctl} and \emph{rsm.ctl1} as the predictors. This simulates a situation such as forecasting electric load from a few different weather models.

To be exactly comparable to the SREF scenarios, we generate 21 scenarios for each run-time from the statistical method. Various metrics are used to measure the efficacy of the statistical method. If the scenario were being used to compute the probability of certain events we might evaluate the scenarios for that express purpose. We, however, wanted to evaluate the scenario forecasting method agnostic to the events that might be of interest.

A good metric to evaluate a scenario forecast would estimate the likelihood of observing the observed target values given the distribution of the scenarios. Because this is difficult to do from a finite sample of scenarios, two different metrics that measure the \emph{calibration} and \emph{sharpness} of the scenarios. Calibration measures whether the target is drawn from the same distribution as the scenarios, and sharpness measures how tightly the scenarios cluster around the target (see~\cite{Gneting_cal_sharp} for a detailed discussion of the desiderata for probabilistic forecasts).

\begin{figure}[h]
\centering
\includegraphics[width=0.8\textwidth]{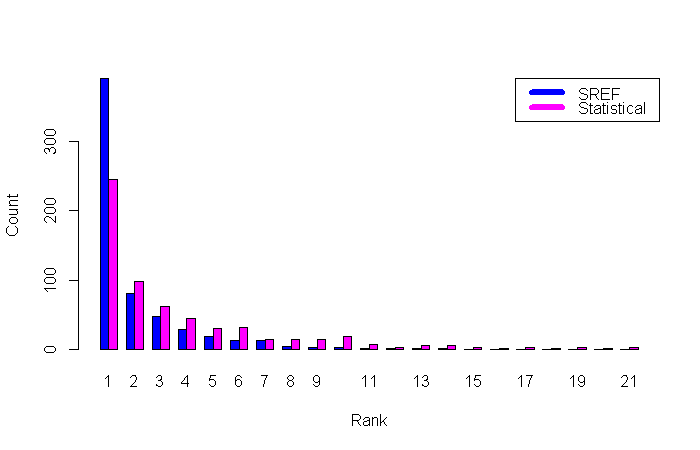}
\caption{Minimum spanning tree rank histogram.}
\label{fig:msp}
\end{figure}

One metric used to judge the calibration for univariate scenarios is the rank histogram, which measures the flatness of the histogram of the ranks of the actual targets among the scenarios. This is difficult to generalize to multivariate targets (as is our lot). For low dimensional problems, one method is the \emph{minimum spanning tree distance rank histogram}~\cite{msp_rankhistogram}. Here the length of the Euclidean minimum spanning tree is computed for the scenarios, as well as for each graph where one of the scenarios is replaced with the actual target. The flatness of the rank of the first length among all the lengths is a measure of how well-calibrated the scenarios are. Figure~\ref{fig:msp} shows the MSP rank histograms for the SREF and the statistical method. It appears that neither is very well calibrated even if our statistical method is an improvement. (It is possible that the MSP rank histogram method is not suitable for such high dimensional problems as ours.)

We might expect from a scenario forecasting system that at least one of the possible scenarios that are forecast ends up being close to the truth. Therefore we may compare two scenario generation methods by comparing the \emph{closest} scenarios to the actual target. (Clearly this makes sense only when the two methods produce the same number of scenarios.)

\begin{figure}[h]
\centering
\includegraphics[width=0.6\textwidth]{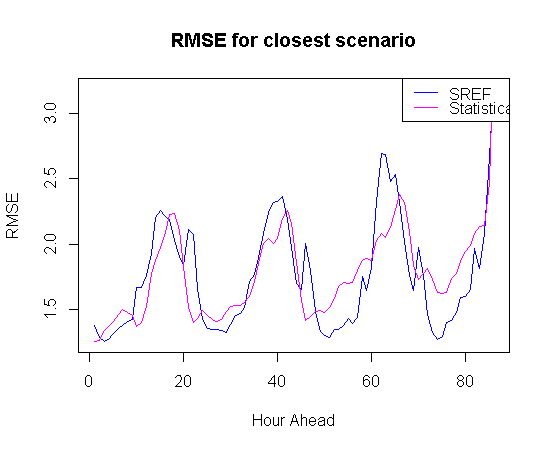}
\caption{RMSE of the closest scenario by forecast horizon.}
\label{fig:rmse}
\end{figure}

\begin{figure}[h]
\centering
\includegraphics[width=0.6\textwidth]{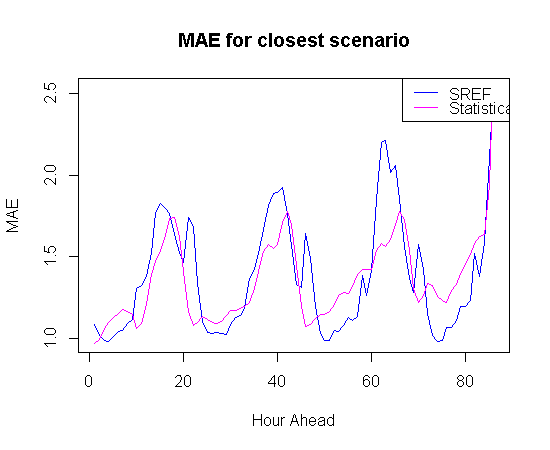}
\caption{MAE of the closest scenario by forecast horizon.}
\label{fig:mae}
\end{figure}

\begin{figure}[h]
\centering
\includegraphics[width=0.6\textwidth]{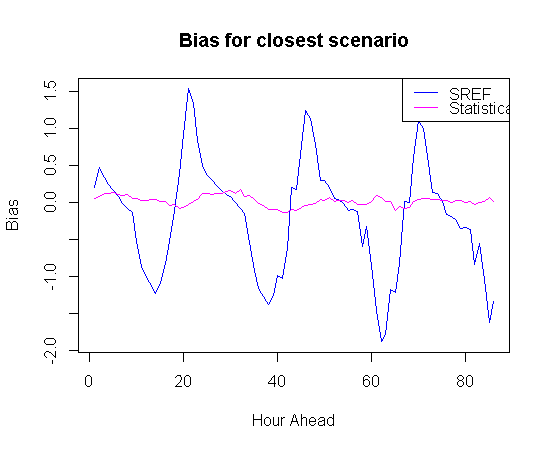}
\caption{Bias of the closest scenario by forecast horizon.}
\label{fig:bias}
\end{figure}

Figures \ref{fig:rmse}, \ref{fig:mae} and \ref{fig:bias} show the RMSE, MAE and Bias of the closest scenario respectively, as function of the forecast horizon. It appears that the SREF models tend to over-forecast the temperature at night and under forecast at midday. 

\begin{figure}[h]
\centering
\includegraphics[width=0.7\textwidth]{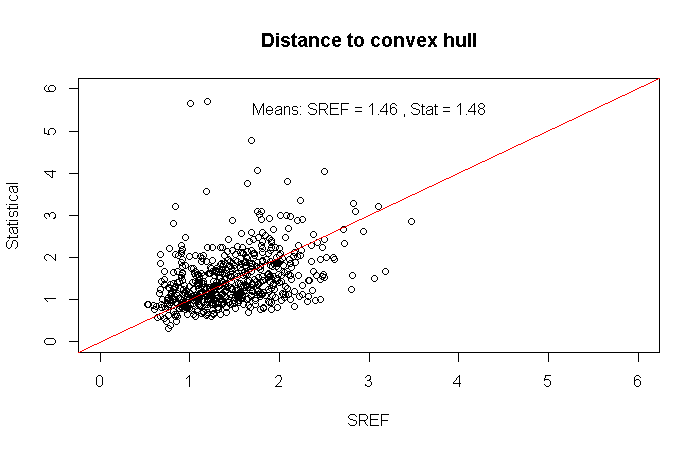}
\caption{Distance of the actual time-series to the convex hull of the scenarios.}
\label{fig:cvxhull}
\end{figure}

\begin{figure}[htb]
\centering
\includegraphics[width=0.9\textwidth]{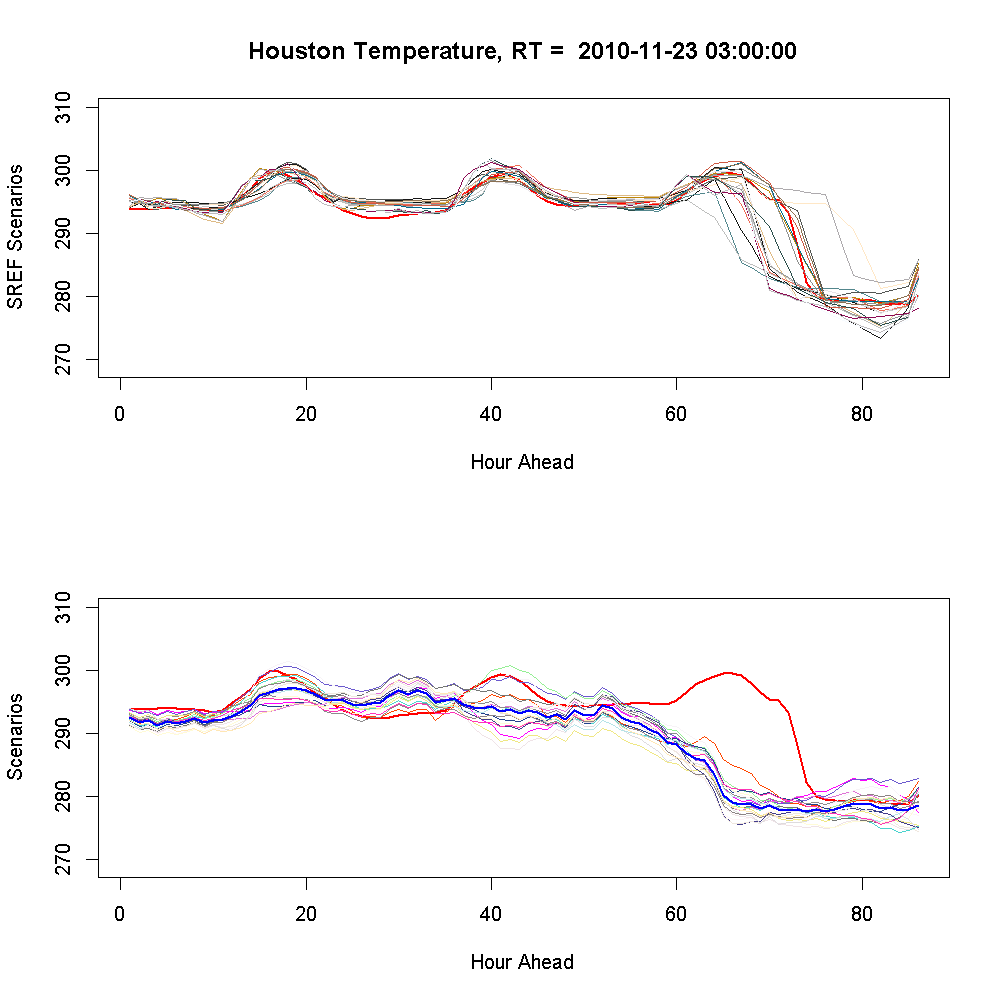}
\caption{Top: SREF, Bottom: Statistical. The red line is the actual in both plots and the blue line in the bottom plot is the mean of the ensemble.}
\label{fig:ts12}
\end{figure}

Figure~\ref{fig:cvxhull} shows the scatter plot of the normalized Euclidean distance of the actual target to the convex hull of the scenarios for the statistical method against that of SREF. The statistical method is slightly worse owing to a few instances where the actual temperature time-series were far from all the scenarios (see Figure~\ref{fig:ts12}).

\begin{figure}[h]
\centering
\includegraphics[width=0.7\textwidth]{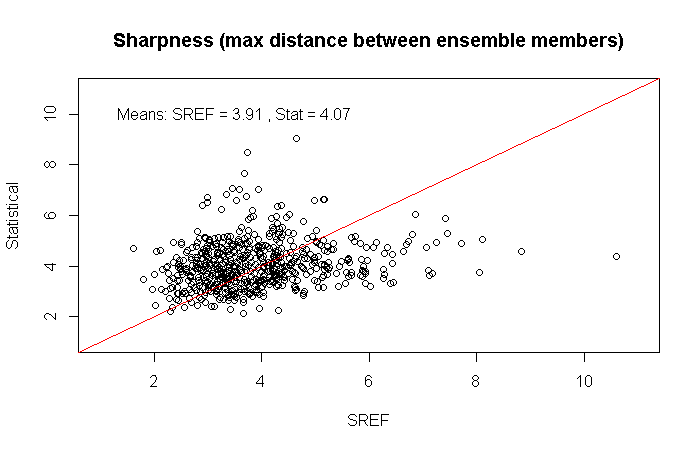}
\caption{Sharpness SREF vs statistical.}
\label{fig:sharpness}
\end{figure}

We measure the sharpness of the scenario forecasts by the Euclidean distance between the two furthest members of the ensemble. Figure~\ref{fig:sharpness} shows the scatter plot of the sharpness of the statistical method against the SREF for each of the approximately 600 instances. We note that the statistical ensemble is in general less sharp than the SREF. There is a trade-off between sharpness and calibration, and we expect that with a larger training set we can improve sharpness for a given calibration.

\begin{figure}[htb]
\centering
\includegraphics[width=0.9\textwidth]{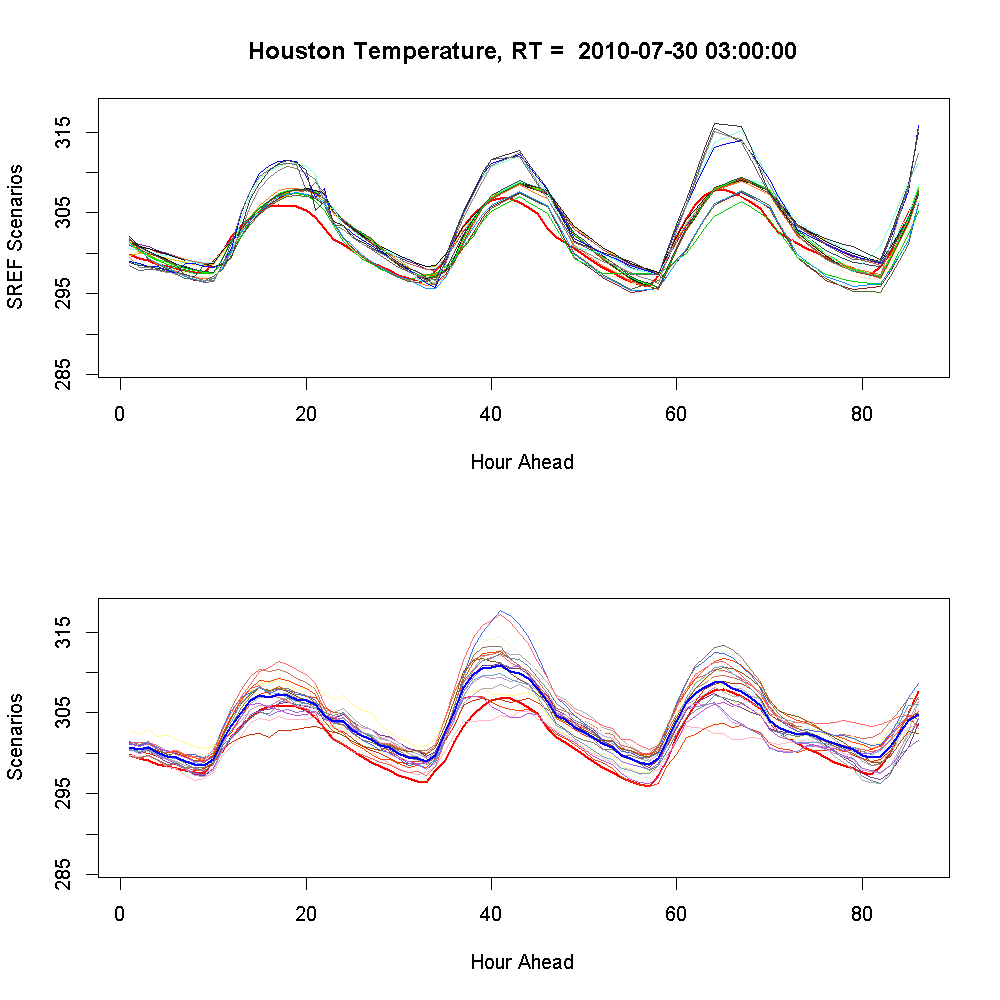}
\caption{Top: SREF, Bottom: Statistical. The red line is the actual in both plots and the blue line in the bottom plot is the mean of the ensemble.}
\label{fig:ts1}
\end{figure}

\begin{figure}[htb]
\centering
\includegraphics[width=0.9\textwidth]{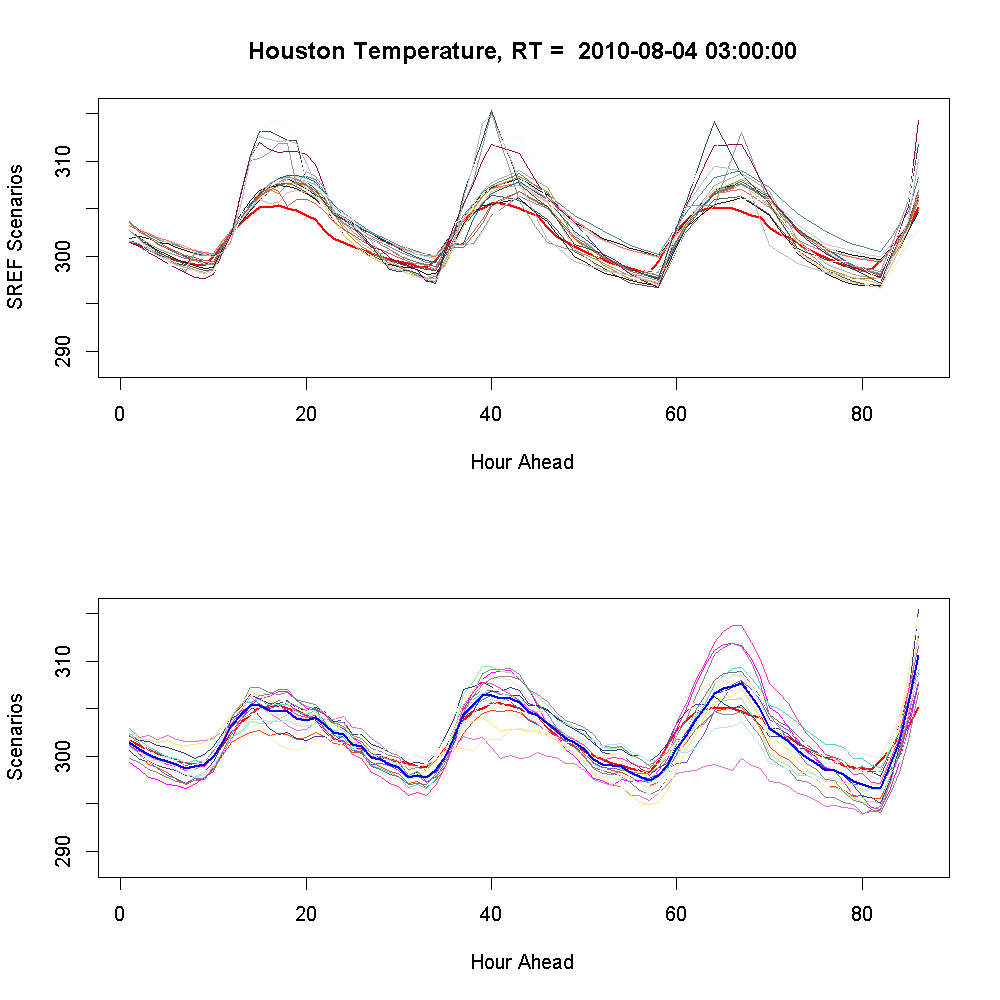}
\caption{Top: SREF, Bottom: Statistical. The red line is the actual in both plots and the blue line in the bottom plot is the mean of the ensemble.}
\label{fig:ts2}
\end{figure}

\begin{figure}[htb]
\centering
\includegraphics[width=0.9\textwidth]{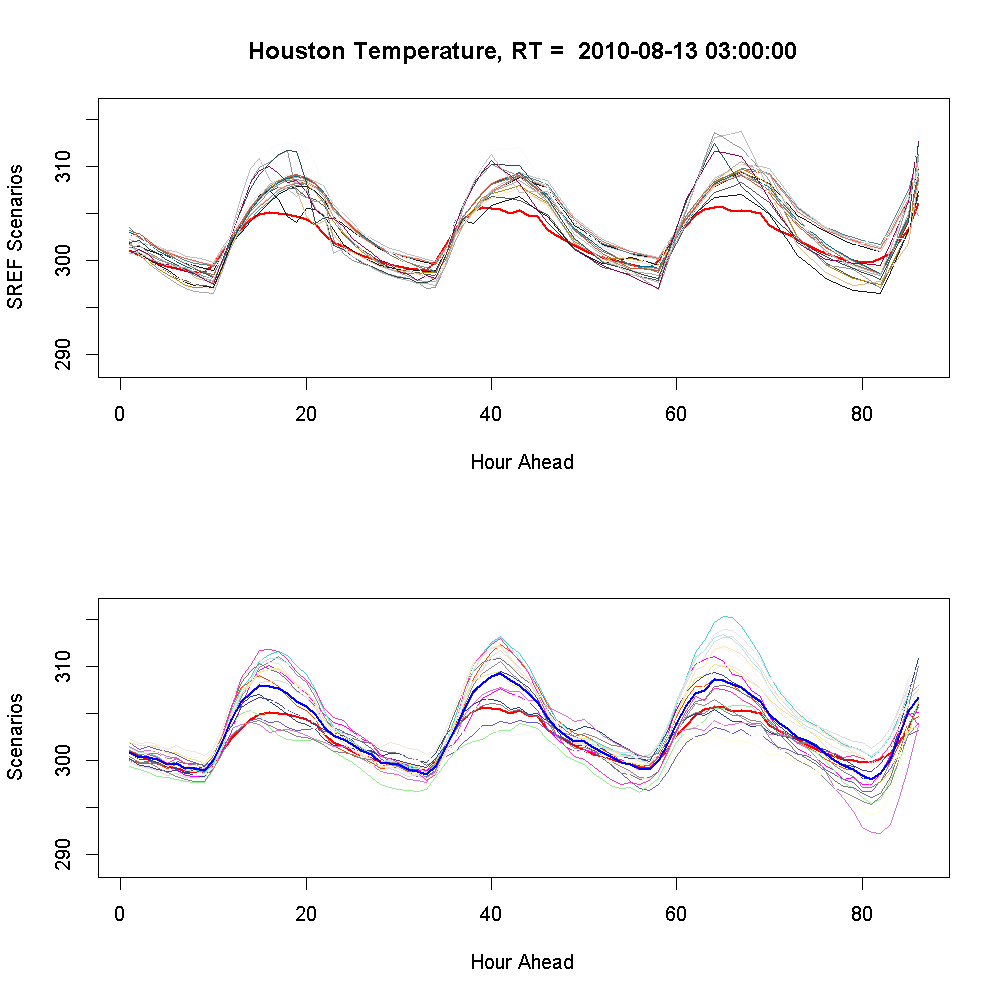}
\caption{Top: SREF, Bottom: Statistical. The red line is the actual in both plots and the blue line in the bottom plot is the mean of the ensemble.}
\label{fig:ts3}
\end{figure}

\begin{figure}[htb]
\centering
\includegraphics[width=0.9\textwidth]{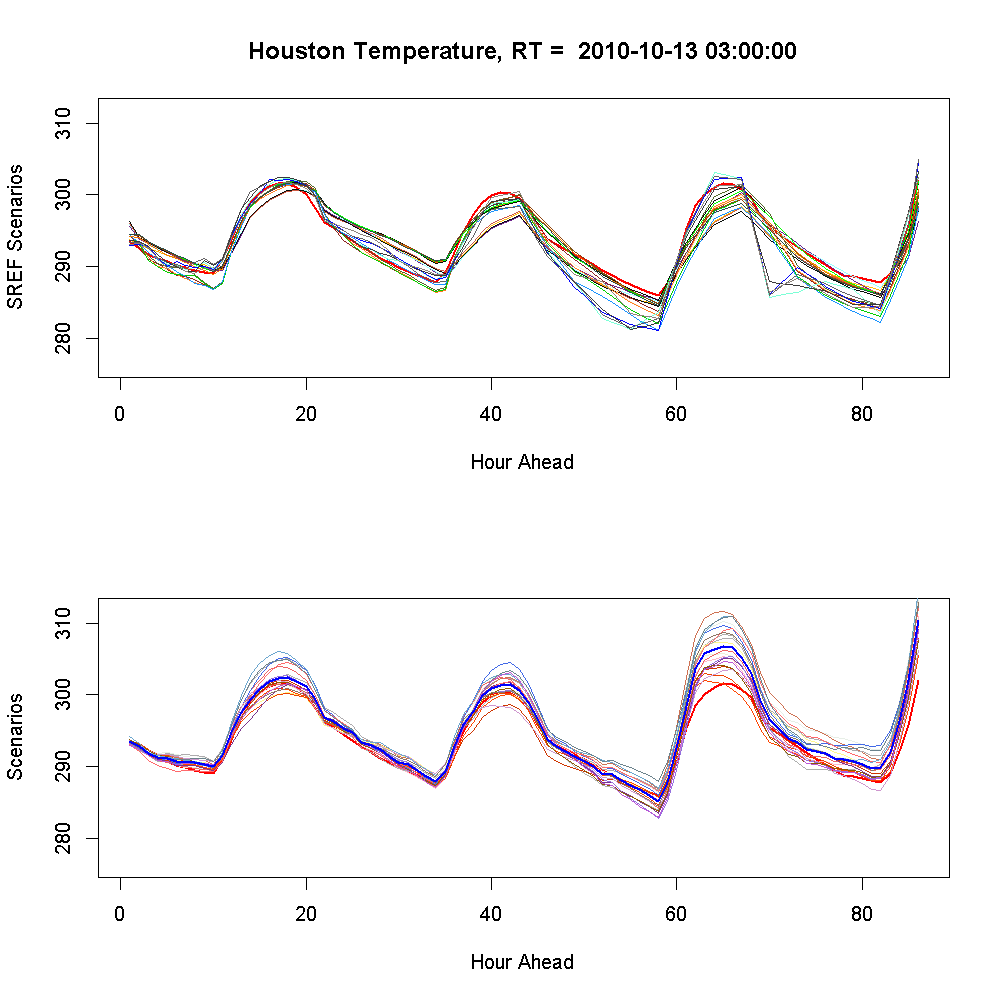}
\caption{Top: SREF, Bottom: Statistical. The red line is the actual in both plots and the blue line in the bottom plot is the mean of the ensemble.}
\label{fig:ts8}
\end{figure}

\begin{figure}[htb]
\centering
\includegraphics[width=0.9\textwidth]{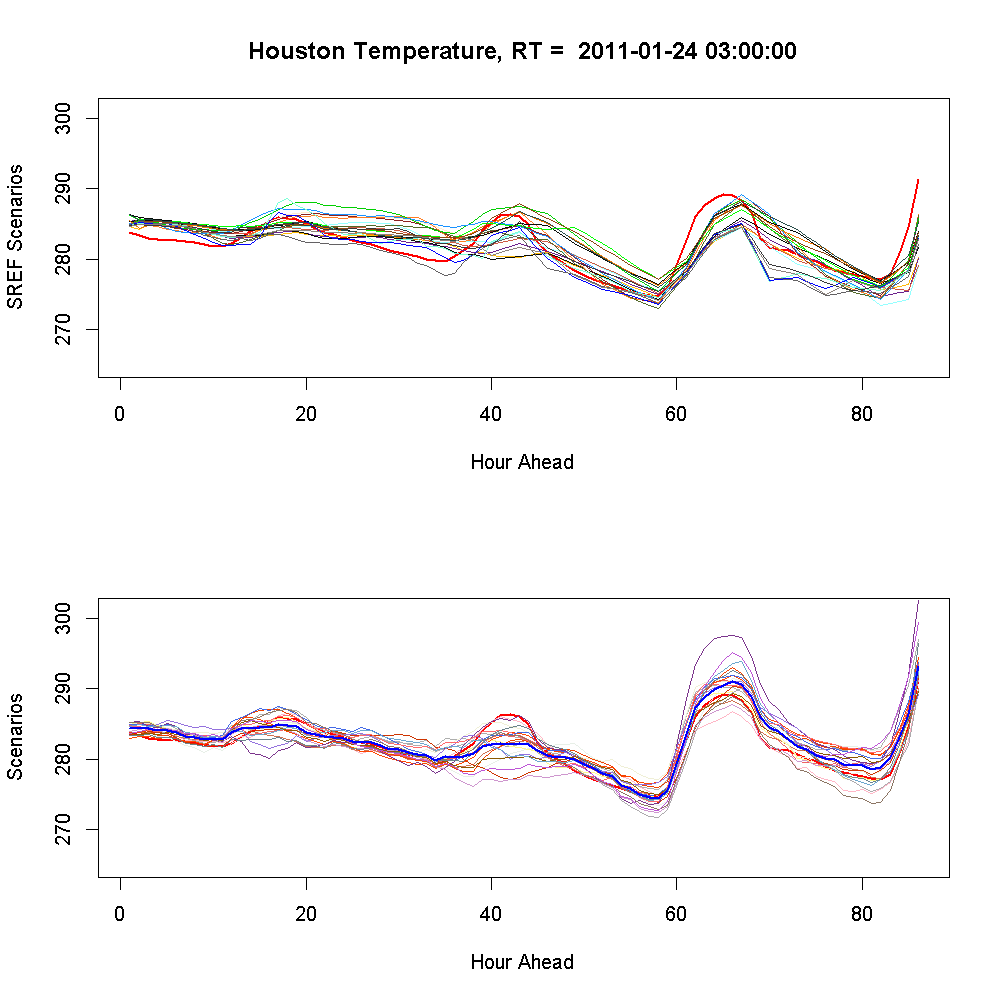}
\caption{Top: SREF, Bottom: Statistical. The red line is the actual in both plots and the blue line in the bottom plot is the mean of the ensemble.}
\label{fig:ts15}
\end{figure}

\begin{figure}[htb]
\centering
\includegraphics[width=0.9\textwidth]{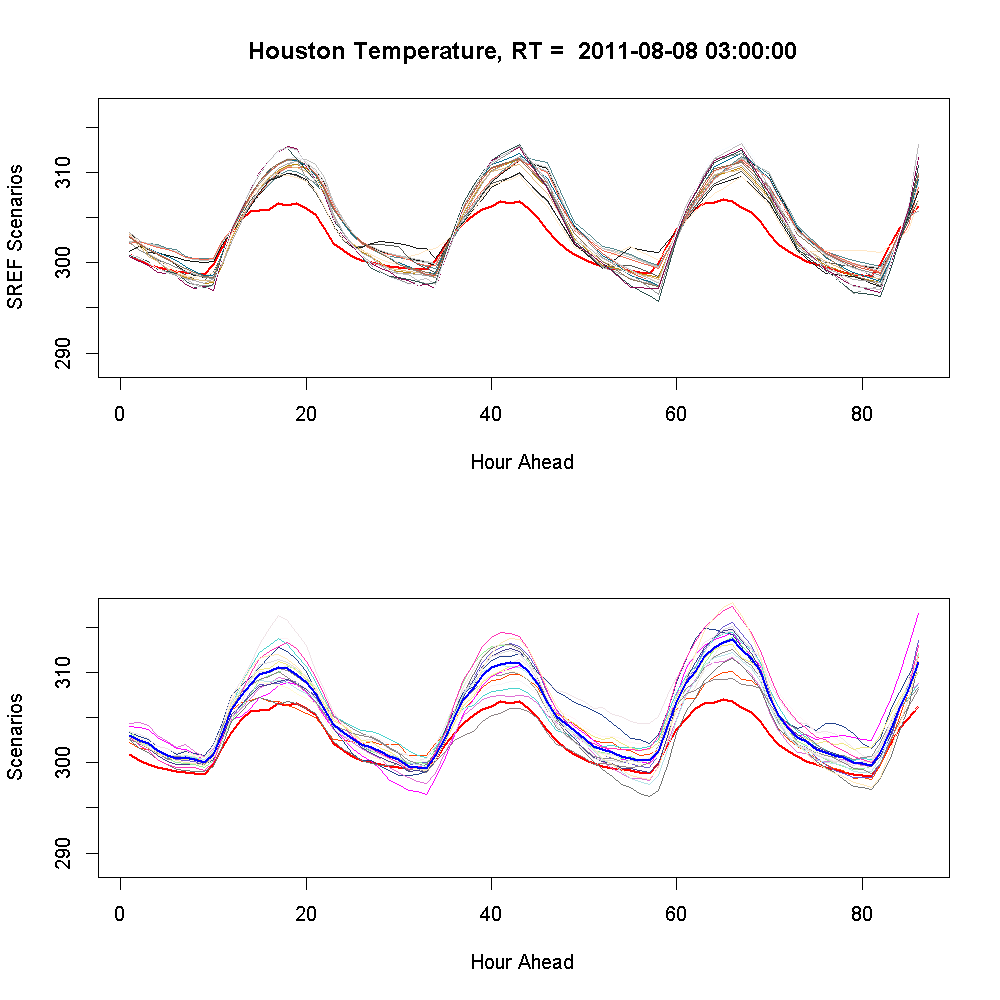}
\caption{Top: SREF, Bottom: Statistical. The red line is the actual in both plots and the blue line in the bottom plot is the mean of the ensemble.}
\label{fig:ts19}
\end{figure}

\begin{figure}[htb]
\centering
\includegraphics[width=0.9\textwidth]{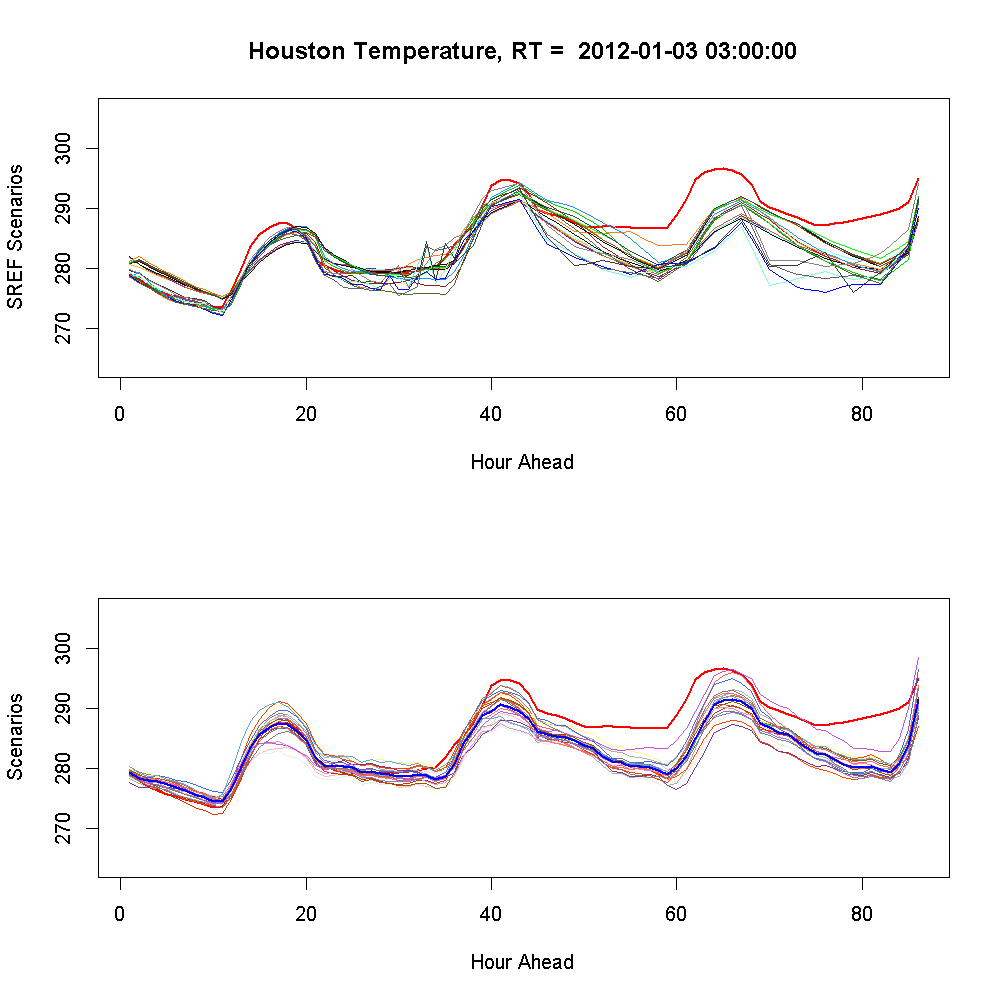}
\caption{Top: SREF, Bottom: Statistical. The red line is the actual in both plots and the blue line in the bottom plot is the mean of the ensemble.}
\label{fig:ts23}
\end{figure}

\begin{figure}[htb]
\centering
\includegraphics[width=0.9\textwidth]{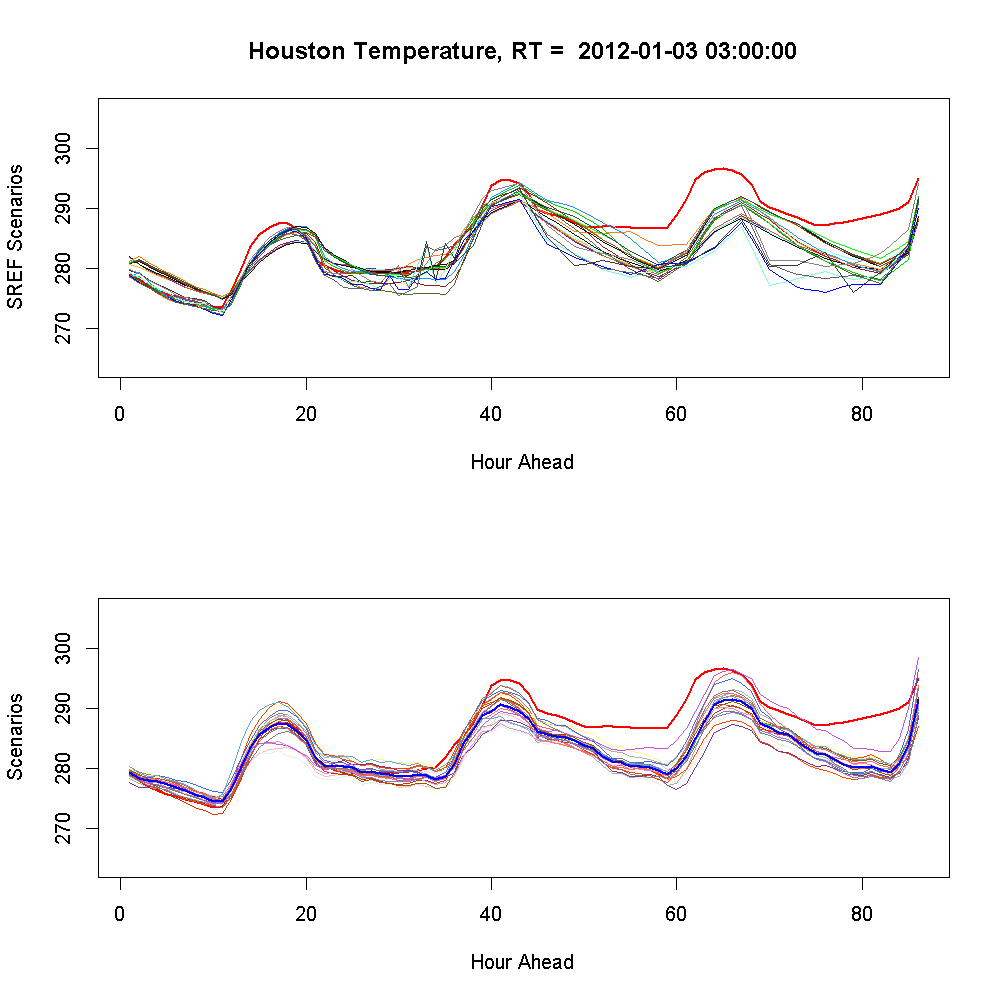}
\caption{Top: SREF, Bottom: Statistical. The red line is the actual in both plots and the blue line in the bottom plot is the mean of the ensemble.}
\label{fig:tsend}
\end{figure}

Figures \ref{fig:ts1} through \ref{fig:tsend} show the scenarios generated by SREF and the statistical method for various run-times. 

\section{Conclusion and Future Work}
The preliminary experimental results presented above show that the proposed method holds promise for time-series scenario forecasting. Although our statistical approach performs about the same or perhaps slightly worse than the physics-based method, it has the advantages that it can be better calibrated, enables a computationally cheap way to generate a large number of scenarios, and can be applied to variables for which physical models are absent. The main disadvantages are that, being a statistical approach, it is limited by the scenarios observed in the training data, and the possibility that the scenarios violate known physical constraints. 

Avenues for future work include
\begin{itemize}
%\item Applying the method for forecasting wind power, electric load, locational marginal prices etc.
\item Scenario forecasting a combination of two or more weather variables.
\item Experimenting with scenarios over patches of time-series (i.e., sliding windows) to exploit self similarity. Such methods have shown promise in image de-noising and interpolation. 
\item Modifying the hierarchical model to incorporate prior knowledge about temporal and spatial correlations.
\item Studying the impact of training data size on learning high dimensional dictionaries for scenario forecasting. 
\end{itemize}

%\begin{figure}[htb]
%\centering
%\includegraphics[width=0.9\textwidth]{plots/plot_k_100_31.png}
%\caption{Top: SREF, Bottom: Statistical. The red line is the actual in both plots and the blue line in the bottom plot is the mean of the ensemble.}
%\label{fig:ts4}
%\end{figure}

%\begin{figure}[htb]
%\centering
%\includegraphics[width=0.9\textwidth]{plots/plot_k_100_53.png}
%\caption{Top: SREF, Bottom: Statistical. The red line is the actual in both plots and the blue line in the bottom plot is the mean of the ensemble.}
%\label{fig:ts5}
%\end{figure}
%
%
%\begin{figure}[htb]
%\centering
%\includegraphics[width=0.9\textwidth]{plots/plot_k_100_55.png}
%\caption{Top: SREF, Bottom: Statistical. The red line is the actual in both plots and the blue line in the bottom plot is the mean of the ensemble.}
%\label{fig:ts6}
%\end{figure}
%
%
%\begin{figure}[htb]
%\centering
%\includegraphics[width=0.9\textwidth]{plots/plot_k_100_68.png}
%\caption{Top: SREF, Bottom: Statistical. The red line is the actual in both plots and the blue line in the bottom plot is the mean of the ensemble.}
%\label{fig:ts7}
%\end{figure}

%\begin{figure}[htb]
%\centering
%\includegraphics[width=0.9\textwidth]{plots/plot_k_100_79.png}
%\caption{Top: SREF, Bottom: Statistical. The red line is the actual in both plots and the blue line in the bottom plot is the mean of the ensemble.}
%\label{fig:ts9}
%\end{figure}
%
%
%\begin{figure}[htb]
%\centering
%\includegraphics[width=0.9\textwidth]{plots/plot_k_100_90.png}
%\caption{Top: SREF, Bottom: Statistical. The red line is the actual in both plots and the blue line in the bottom plot is the mean of the ensemble.}
%\label{fig:ts10}
%\end{figure}
%
%
%\begin{figure}[htb]
%\centering
%\includegraphics[width=0.9\textwidth]{plots/plot_k_100_105.png}
%\caption{Top: SREF, Bottom: Statistical. The red line is the actual in both plots and the blue line in the bottom plot is the mean of the ensemble.}
%\label{fig:ts11}
%\end{figure}

%\clearpage
%\begin{figure}[htb]
%\centering
%\includegraphics[width=0.9\textwidth]{plots/plot_k_100_124.png}
%\caption{Top: SREF, Bottom: Statistical. The red line is the actual in both plots and the blue line in the bottom plot is the mean of the ensemble.}
%\label{fig:ts13}
%\end{figure}
%
%
%\begin{figure}[htb]
%\centering
%\includegraphics[width=0.9\textwidth]{plots/plot_k_100_148.png}
%\caption{Top: SREF, Bottom: Statistical. The red line is the actual in both plots and the blue line in the bottom plot is the mean of the ensemble.}
%\label{fig:ts14}
%\end{figure}

%\begin{figure}[htb]
%\centering
%\includegraphics[width=0.9\textwidth]{plots/plot_k_100_187.png}
%\caption{Top: SREF, Bottom: Statistical. The red line is the actual in both plots and the blue line in the bottom plot is the mean of the ensemble.}
%\label{fig:ts16}
%\end{figure}
%
%
%\begin{figure}[htb]
%\centering
%\includegraphics[width=0.9\textwidth]{plots/plot_k_100_187.png}
%\caption{Top: SREF, Bottom: Statistical. The red line is the actual in both plots and the blue line in the bottom plot is the mean of the ensemble.}
%\label{fig:ts17}
%\end{figure}
%
%
%\begin{figure}[htb]
%\centering
%\includegraphics[width=0.9\textwidth]{plots/plot_k_100_199.png}
%\caption{Top: SREF, Bottom: Statistical. The red line is the actual in both plots and the blue line in the bottom plot is the mean of the ensemble.}
%\label{fig:ts18}
%\end{figure}

\clearpage

%\begin{figure}[htb]
%\centering
%\includegraphics[width=0.9\textwidth]{plots/plot_k_100_371.png}
%\caption{Top: SREF, Bottom: Statistical. The red line is the actual in both plots and the blue line in the bottom plot is the mean of the ensemble.}
%\label{fig:ts20}
%\end{figure}
%
%
%\begin{figure}[htb]
%\centering
%\includegraphics[width=0.9\textwidth]{plots/plot_k_100_414.png}
%\caption{Top: SREF, Bottom: Statistical. The red line is the actual in both plots and the blue line in the bottom plot is the mean of the ensemble.}
%\label{fig:ts21}
%\end{figure}
%
%
%\begin{figure}[htb]
%\centering
%\includegraphics[width=0.9\textwidth]{plots/plot_k_100_476.png}
%\caption{Top: SREF, Bottom: Statistical. The red line is the actual in both plots and the blue line in the bottom plot is the mean of the ensemble.}
%\label{fig:ts22}
%\end{figure}

%\begin{figure}[htb]
%\centering
%\includegraphics[width=0.9\textwidth]{plots/plot_k_100_501.png}
%\caption{Top: SREF, Bottom: Statistical. The red line is the actual in both plots and the blue line in the bottom plot is the mean of the ensemble.}
%\label{fig:ts24}
%\end{figure}
%
%\begin{figure}[htb]
%\centering
%\includegraphics[width=0.9\textwidth]{plots/plot_k_100_559.png}
%\caption{Top: SREF, Bottom: Statistical. The red line is the actual in both plots and the blue line in the bottom plot is the mean of the ensemble.}
%\label{fig:ts25}
%\end{figure}
%
%\begin{figure}[htb]
%\centering
%\includegraphics[width=0.9\textwidth]{plots/plot_k_100_576.png}
%\caption{Top: SREF, Bottom: Statistical. The red line is the actual in both plots and the blue line in the bottom plot is the mean of the ensemble.}
%\label{fig:ts26}
%\end{figure}

\end{document}